\def\year{2019}\relax
\documentclass[letterpaper]{article} %DO NOT CHANGE THIS
\usepackage{aaai19}  %Required
\usepackage{times}  %Required
\usepackage{helvet}  %Required
\usepackage{courier}  %Required
\usepackage{url}  %Required
\usepackage{graphicx}  %Required
\begin{document}
\title{Evolutionarily-Curated Curriculum Learning\\ for Deep Reinforcement Learning Agents
}

%\author{Anonymous}
\author{Michael Cerny Green, Benjamin Sergent, Pushyami Shandilya, Vibhor Kumar\\Imbellus, Los Angeles, CA}

\maketitle

%%%%%%%%%%%%%%%%%%%%%%%%%%%%%%%%%%%%%%%%%%%%%%%%%%%%%%%%%%%%%%%%%%%%%%%%%%%%%%%%
\begin{abstract}

In this paper we propose a new training loop for deep reinforcement learning agents with an evolutionary generator. Evolutionary procedural content generation has been used in the creation of maps and levels for games before. Our system incorporates an evolutionary map generator to construct a training curriculum that is evolved to maximize loss within the state-of-the-art Double Dueling Deep Q Network architecture with prioritized replay~\cite{wang2016dueling}~\cite{schaul2015prioritized}. We present a case-study in which we prove the efficacy of our new method on a game with a discrete, large action space we made called \emph{Attackers and Defenders}. Our results demonstrate that training on an evolutionarily-curated curriculum (directed sampling) of maps both expedites training and improves generalization when compared to a network trained on an undirected sampling of maps.
\end{abstract}

%%%%%%%%%%%%%%%%%%%%%%%%%%%%%%%%%%%%%%%%%%%%%%%%%%%%%%%%%%%%%%%%%%%%%%%%%%%%%%%%
\section{Introduction}
% TODO vkumar or mgreen: We need to rewrite this, or use the next paragraph to start.
The use of games as benchmarks for AI progress has propagated to nearly the entire AI research community, including Chess, Atari Breakout, and more recently with Go as superhuman agents have been developed. Many recent papers document new AI methods being used within game environments. But the current state-of-the-art training method to ensure generalization in a neural network system of any kind remains brute force random sampling training, i.e. give a network enough unique states to train on and hope generalization naturally occurs. We propose a new method of directed sampling training called 'evolutionarily-curated curriculum learning' (ECCL), which we argue results in faster and better network generalization.

Past experiments have shown the potential in teaching simple concepts first, on which more complicated ones can then be taught, as a successful way to train networks~\cite{elman1993learning}. To go one step further, we propose a method which specifically identifies weaknesses in the network and then generates content that force the network to face these weaknesses head-on. Our system dynamically evolves a curriculum by searching for content that maximizes the network's loss, which makes the network generalize faster and perform better.

In this paper, we give a brief overview of research within reinforcement learning, the use of evolutionary algorithms in procedural content generation, and curriculum learning research for networks in Section \ref{sec:background}. In Section \ref{sec:theory} we discuss the theory of evolutionarily-based curriculum learning and how it could be applied to a reinforcement learning agent. We then use \emph{Attackers and Defenders} as a case study in Section \ref{sec:case}, with results and discussion from our experiment in Section \ref{sec:results/discussion}, and conclude in Section \ref{sec:conclusion}.

\begin{figure}[tp]
\begin{center}
\includegraphics[width=\linewidth]{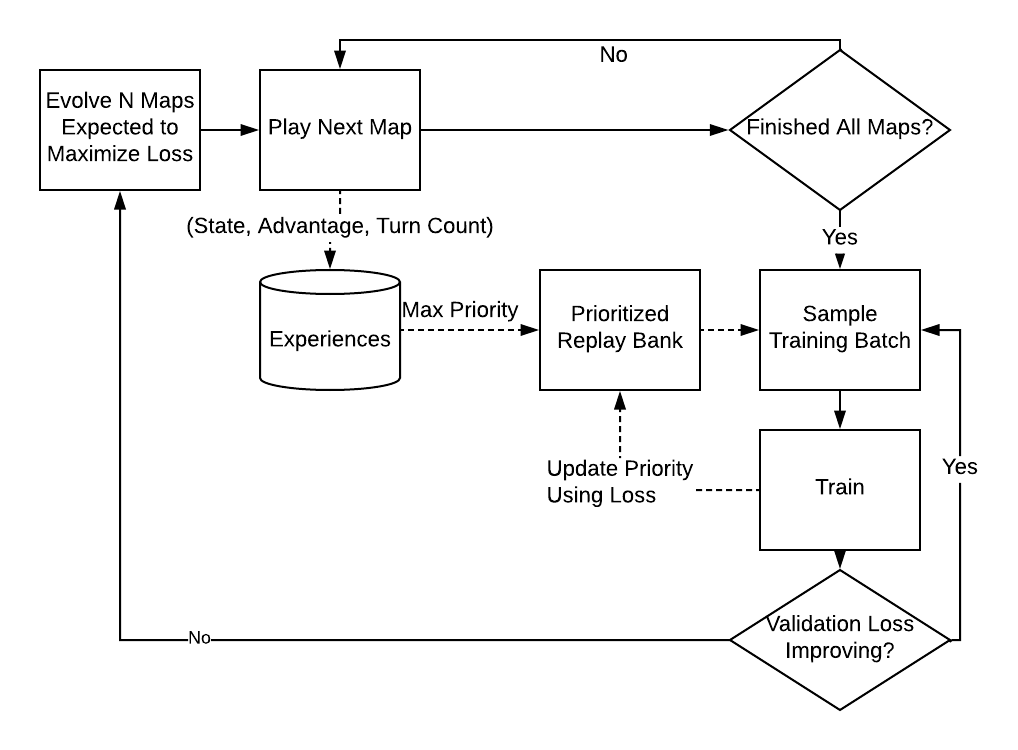}
\caption{Evolutionarily-based curriculum learning in an agent's training loop}
\label{fig:training-loop}
\end{center}
\end{figure}

\section{Background}\label{sec:background}
This section begins with a brief overview of deep reinforcement learning research, beginning with Minksy in 1954~\cite{minsky1954theory} and finishing with the state-of-the-art DDDQN~\cite{wang2016dueling}, AlphaZero~\cite{silver2017mastering,silver2017masteringchess}, and ExIt~\cite{anthony2017thinking} agent architectures. It then discusses evolutionary algorithms and how they can be applied toward procedural content generation in games. The section concludes with the concept of curriculum learning for machines and the admittedly scant amount of research within this area.

\subsection{Deep Reinforcement Learning}
Reinforcement Learning (RL) concerns itself with the idea of learning through trial-and-error interactions with a dynamic environment and balancing the reward trade-off between long-term and short-term planning~\cite{sutton1998introduction}. 
RL has been studied since Minksy \cite{minsky1954theory} in the 1950's. 
% Sutton 1984, 1988 - temporal difference learning method
% Barto, sutton and anderson 1983 - actor critic algorithms
% Watkins 1989 - Q learning
Since then, important improvements to the concept have been advanced including the temporal difference learning method~\cite{sutton1984temporal,sutton1988learning}, on which q-learning~\cite{watkins1992q} and actor-critic~\cite{barto1983neuronlike} techniques are built. 
% Williams 1992 and Gullapalli 1990 - ANNs using RL
Gullapalli~\cite{gullapalli1990stochastic} and Williams~\cite{williams1992simple} are early examples of the use of RL within artificial neural networks (ANNs). 
% Rumelhart et al. 1998 - boom currently fueled by Backprop
%  - also mention advances in GPU and CPUs contributing to this
When Rumelhart et al discovered the \emph{backpropogation} 
algorithm~\cite{rumelhart1986learning}, deep learning took off in popularity. This has been bolstered recently by the rise in the capability and affordability of computer-processing units and graphical-processing units. 
% Kaelbling Littman, and Moore 1996, Sutton and Barto 1998, Szepesv\'ari 2010 - in-depth overview of RL
Further readings of work in RL can be found in reviews by Schmidhuber~\cite{schmidhuber2015deep} and Szepsv\'ari~\cite{szepesvari2010algorithms}.

RL applied to deep learning has been only recently successful due to some key advancements. 
% Mnih et al 2015 - Deep-Q-Networks (DQN) - experience replay and target networks to stabilize learning
Mnih et al proposed Deep Q Networks (DQNs)~\cite{mnih2015human} using target networks and experience replay to improve the known divergence issues present in RL. 
% - extensions to DQN = Double Deep QN (van Hesselt, Guez, and Silver 2016), Dueling network architecture (Wang et al 2016)
van Hesselt et al built Double Deep Q Networks~\cite{van2016deep} which help reduce the overestimation errors that normal DQNs suffer from. Hessel et al wrote a paper surveying state-of-the-art improvements to DQN within the Atari framework, including Double Dueling Deep Q Networks, Distributional Deep Q Networks, and Noisy Deep Q Networks~\cite{hessel2017rainbow}.
% Schaul et al 2015 - Prioritized experience replay 
Prioritized experience reply~\cite{schaul2015prioritized} allows DQNs to remember past experiences in a prioritized fashion, rather than at the same frequency that they were experienced.

% Dueling networks by Wang et al 2016
Further DQN stability has been attained through the introduction of dueling networks~\cite{wang2016dueling}, which are built on top of Double DQNs. Dueling networks use two separate estimators for the state value function and state-dependent action advantage function to generalize learning across actions without effecting the underlying RL algorithm. Double Dueling DQNs are currently considered state-of-the-art reinforcement learning algorithms.

% Mnih et al 2016 - AC3, Asynchronous Advantage Actor Critic model
Asynchronous Advantage Actor Critic (A3C) networks were created by Mnih et al in a successful attempt to apply neural networks to actor-critic reinforcement learning. The ``asynchronous'' part of A3C makes training parallelizable, allowing for massive computation speedups.

% alpha go
The AlphaGo algorithm combined Monte Carlo Tree Search (MCTS) with deep neural networks to play the game of Go and become the first AI to beat the human world champion of the game~\cite{silver2016mastering}. 
% alphazero in go and chess
The more advanced version, AlphaZero, was able to learn only by self-playing and outperformed AlphaGo~\cite{silver2017mastering}. The same AlphaZero architecture was then applied to both Chess and Shogi to convincingly beat world-champion programs~\cite{silver2017masteringchess}. 
% Anthony et al 2017 - Expert Iteration algorithm
At the same time, Anthony et al discovered the Expert Iteration algorithm~\cite{anthony2017thinking}, which also uses a neural network policy to guide tree search. Since the advent of these state-of-the-art algorithms, much research has been done to either improve or apply them to different problems with varying degrees of success.

\subsection{Evolution and Procedural Content Generation}
Evolutionary algorithms (EA) fall within the area of optimization search inspired by Darwinian evolutionary concepts such as reproduction, fitness, and mutation \cite{togelius2016introduction}. EA has been used within games to procedurally generate levels, game elements within them, and sometimes even games themselves~\cite{khalifa2017general}. Puzzle generation is a primary example of this kind of search-based generation~\cite{ashlock2010automatic}, which can be used to create puzzles with a desired solution difficulty. \emph{Checkpoint based fitness} allows for fitness function parameterization~\cite{ashlock2011search}, affording substantial control over generated properties. Stylistic generation is made possible by using fashion-based cellular automata~\cite{ashlock2015evolvable}. An EA generator for a given game can also evolve many things at once by decomposing level generation into multiple parts. McGuinness et al did this by creating a micro evolutionary system which evolves individual tile sections of a level and an overall macro generation system which evolves placement patterns for the tiles~\cite{mcguinness2011decomposing}. 
Evolutionary search can be used for generalized level generation in multiple domains such as General Video Game AI~\cite{khalifa2016general} and PuzzleScript~\cite{khalifa2015automatic}. In later work by Khalifa et al.~\cite{khalifa2018talakat}, they worked on generating levels for a specific game genre (Bullet Hell genre) using a new hybrid evolutionary search called Constrained Map-Elites. The levels were generated using automated playing agents with different parameters to mimic various human play-styles. Green et al. used EA to evolve \emph{Super Mario Bros} (Nintendo, 1985) scenes which taught specific mechanics to the player~\cite{green2018generating}.
We recommend Khalifa's review on searched-based level generation for further reading into EA for generation in games~\cite{khalifa2015literature}.

\subsection{Curriculum Learning in Machines}
The concept of curriculum learning (CL) in machines can be traced back to Elman in 1993~\cite{elman1993learning}. The basic idea is to keep initial training data simple and slowly ramp up in difficulty as the model learned.  Krueger and Dayan~\cite{krueger2009flexible} did a cognitive-based analysis with evidence that shaping data provided faster convergence. CL was further explored by Bengio et al. in 2009, in an attempt to define several machine learning guided training strategies~\cite{bengio2009curriculum}. Their experiments suggested that incorporating CL into training a model could both speed up training and significantly increase generalization. Recently, Curriculum Learning within adversarial network training~\cite{cai2018curriculum} was explored by Cai et al in an attempt to mitigate ``forgetfulness`` and increase generalization to reduce the effectiveness of adversarial network attacks. 

\subsection{Evolution within Networks}
Using evolutionary strategy for neural networks is a well-researched topic. Genetic algorithms were used in the node weight balancing of a network by Ronald et al~\cite{ronald1994genetic} to evolve a controller for soft-landing a toy lunar module in a simulation. Simultaneously, Gruau evolved the structures and parameters of networks using cellular encoding~\cite{gruau1994neural}. Cartesian Genetic Programming was first designed by Miller et al~\cite{miller1997designing} to design digital circuits using genetic programming. It is called `Cartesian' because of the way it represents a program using a 2-dimensional set of nodes. 

All of these methods and more may be housed under the umbrella of ``neuroevolution'' which is well-defined by Floreano et al~\cite{floreano2008neuroevolution}. A survey of neuroevolution within games is written by Risi and Togelius~\cite{risi2017neuroevolution}. We mention neuroevolution to highlight the major difference that whereas it is used to evolve the parameters or architecture of a network, our approach evolves training data as part of a curriculum for a constant architecture.

\begin{figure}
\includegraphics[width=\linewidth]{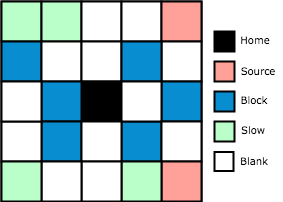}
\caption{A visualization of a map in \textit{Attackers and Defenders}}
\label{fig:map-rep}
\end{figure}

\section{Evolved Curriculum in the Training Loop}\label{sec:theory}

Traditional training of a neural network involves a training schedule established by taking random batches of a fixed training set, which is assumed to be an unbiased random sample of the data space. For game-playing agents specifically, this training set is the set of levels or maps that the network is exposed to. The hope is that the sample is sufficient to train the neural network to generalize to the entire data space (i.e. all possible maps). 

Instead, ECCL relies on producing a training curriculum composed of a biased sampling of the data space, specifically designed to improve generalization. ECCL involves two parts: the agent to be trained and an evolutionary generator. Unlike neuroevolution methods, the use of evolution in evolved curriculum is \emph{not} to evolve the weights or architecture of the network. The evolutionary generator's sole purpose is to evolve \emph{scenarios} which have the best potential to improve the agent's generalization. To do this, the goal of the evolutionary generator is to produce maps which maximize the loss of the agent network. 

Figure \ref{fig:training-loop} shows the training loop that uses evolutionarily-based curriculum learning. In this figure, the agent is a Double Dueling Deep Q Network (DDDQN) with Prioritized Replay~\cite{wang2016dueling,schaul2015prioritized} which we used in our case study, explained in Section \ref{sec:case}. One reason for using this specific type of network architecture is the choice of our library- Tensorflow. The network used in ecperiments presents in this paper is near state-of-the-art with a few deletions that we deemed unnecessary to use. 

When the network requests more maps to train on, the evolutionary generator is tasked to evolve maps which maximize the amount of loss in the network by querying the loss directly. By maximizing loss, the network is necessarily seeing a valid map in map space that it failed to generalize to properly. Concretely, this could be a edge-case map that where an uncommon move is optimal or requires another strategy altogether from maps the network has already seen. The agent plays this map, as well as any others in the batch.

% we can ensure that the generator will produce a map that the network has not seen before and could benefit most from training on it. The network benefits most from training on maps it hasn't seen before, since this will increase the possibility for full generalization. The agent then plays this map, as well as any others in the batch. 

After finishing a batch, the system divides the different game states caused by moves from all games into experience snapshots, and sorts the experience snapshots by priority to be stored in a prioritized replay bank. It then uses these experiences to train the weights in the network appropriately, and updates priorities in the bank using network loss. The network then asks the generator to produce more maps to repeat the process until training terminates.

\section{Case Study: \textit{Attackers and Defenders}}\label{sec:case}
The following section concerns a case study in which we compare our method's impact against other state-of-the-art algorithms. We hypothesized that the evolutionary generator would create higher quality training data which would more effectively improve network performance, and our experiment attempts to prove this. 

Section \ref{sec:case-game} explains the game of \textit{Attackers and Defenders} a simple tower defense game which we created as a testbed. Section \ref{sec:case-generator} explains the generator and how it produces maps. Section \ref{sec:case-procedure} describes the training/testing methods used to validate our claims.

\subsection{Attackers and Defenders}\label{sec:case-game}
To prove the concept of ECCL, we created a discrete, large action space, tower-defense game called \textit{Attackers and Defenders} as a test-bed. Figure \ref{fig:map-rep} displays a visualization of a game map. The objective of the player in \textit{Attackers and Defenders} is to prevent enemies from reaching their \textit{home} tile for as long as possible. This game is a model of sequential decision making that applies broadly to other game and non-game domains, which makes it an appropriate testbed for our algorithm.

\subsubsection{Game Entities}\label{sec:case-game-tiles}
\textit{Attacker} entities have hit-points (HP), which may vary in number and generally increase over the course of a single play session. To facilitate this survival goal, the player is given \textit{Defender} entities which do damage to \textit{attacker} HP, \textit{slow} tiles which penalize \textit{attacker} movement when traveling through these tiles, and \textit{block} tiles which prohibit \textit{attacker} movement. Table \ref{table:entities} displays all tiles/entities within the game and how they work.

\begin{table}[tbp]
\begin{tabular}{|l|l|}
\hline
\textbf{Game Entity} & \textbf{Description}                                                                                                         \\ \hline
Neutral              & an empty tile with no penalties                                                                                              \\ \hline
Slow                 & a tile makings attackers 2 turns to move                                                                         \\ \hline
Block                & a tile preventing attackers from moving                                                                       \\ \hline
Home                 & the tile attackers are trying to move onto                                                                                   \\ \hline
Source               & the tiles from which attackers spawn                                                                                         \\ \hline
Attacker             & \begin{tabular}[c]{@{}l@{}}automatous entities which are moving toward\\ the home tile\end{tabular}                          \\ \hline
Defender             & \begin{tabular}[c]{@{}l@{}}entities which the player can place; these do\\ damage to all attackers within range\end{tabular} \\ \hline
\end{tabular}
\caption{A table with all entities in the game}
\label{table:entities}
\end{table}

\subsubsection{Game Loop}\label{sec:case-game-loop}
Each turn, the player is prompted to place a \textit{defender}, \textit{slow}, or \textit{block} tile on the game map.  After the player places an entity, the game advances forward one turn. During this period, a \textit{source} tile may spawn an \textit{attacker}, which will then slowly advance toward the \textit{home} tile. If an \textit{attacker} moves into a space within a \textit{defender}'s attack range (which may overlap with other \textit{defenders}), the \textit{attacker} will suffer damage equivalent to the sum of all in-range \textit{defender} damage. If an an \textit{attacker} runs out of HP, it will be destroyed. If an \textit{attacker} manages to move onto the \textit{home} tile, the game will end.

\subsection{Constructive Generator}\label{sec:case-constructive-generator}
In order to appropriately measure the effect of an evolved curriculum versus the impact of increased access to additional training points, we design the constructive generator. With access to a set of underlying parameters of the data, each with a set of acceptable values, and a global set of constraints, there exists a constructive generator that produces unbiased random samples of this data by simply permuting over possible combinations of parameter values, and simply throwing away those combinations that do not satisfy the constraints. Training with such a generator (which we refer to as our "constructive" network) is analogous to the undirected sampling case where the training data is fixed.

In \textit{Attackers and Defenders}, the constructive generator is given the available tile types and where they can be placed (i.e. parameter values and set of possible values), along with the constraints presented in Table \ref{table:constraint-factors}, and simply selects random combinations of tile values, outputting only those combinations that satisfy the constraints and discarding the rest.

\subsection{Evolutionary Generator}\label{sec:case-generator}
Our system uses the Feasible Infeasible 2-Population (FI-2Pop) genetic algorithm~\cite{kimbrough2008feasible} to evolve boards. FI-2Pop is an evolutionary algorithm which uses two populations: a feasible population and an infeasible population. The infeasible population aims at improving infeasible solutions to ``legally-playable'' threshold, when they become feasible and are transfered to the feasible population. The feasible population, on the other hand, aims at improving the quality of feasible chromosomes. If one becomes infeasible, it is then relocated to the infeasible population. After evolving solutions for several generations, the system outputs the board with the highest fitness.

\subsubsection{Chromosomal Representation, Crossover, and Mutation}
\begin{figure}[h]
\begin{center}
\includegraphics[height=0.8\linewidth]{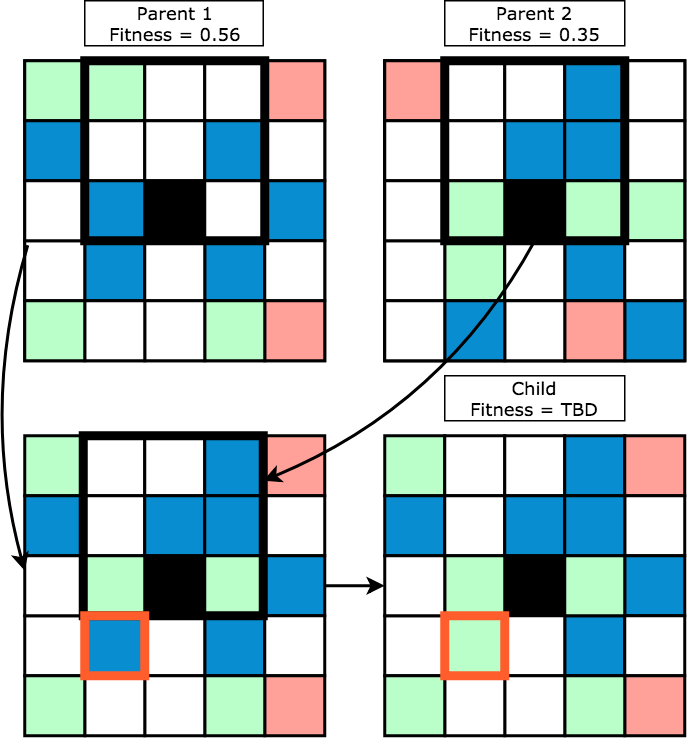}
\caption{The 2D array representation of an \textit{Attackers and Defenders} board. Crossover is shown, using Parent 1 as a template and Parent 2 as a replacement sub-array (black outline). A tile from resulting board is then mutated (red outline), with the fitness of the new child to be calculated later.}
\label{fig:map-crossover}
\end{center}
\end{figure}

A board chromosome is represented as a 2-dimensional array of tile types. Crossover (Figure \ref{fig:map-crossover}) is done using 2-d array crossover, by picking a sub-array within one parent and swapping it with the other, creating a new board as a result. Mutation is done by selecting a random tile and changing its type. Mutation may be performed multiple times on a single board after crossover is completed.

\subsubsection{Evaluating Feasibility and Fitness}
Each board chromosome contains two fitness functions which determine where they fail in terms of feasibility and fitness. The \emph{constrained} fitness dictates whether they are within the infeasible population, and the \emph{feasible} fitness ascertains how optimal of a board it is.

\emph{Constrained fitness} is calculated by averaging the constraint factors listed in Table \ref{table:constraint-factors}. If the constrained average is 1, then this chromosome is feasible. 
\emph{Feasible fitness} is measured by calculating the loss from the agent's loss network on the given map. The larger the loss, the higher the feasible fitness for the chromosome.

\begin{table}[tpb]
\begin{tabular}{|l|l|}
\hline
\textbf{Factor} & \textbf{Description}                                       \\ \hline
Separate Quads  & \% of \textit{sources} in different board quadrants               \\ \hline
Home Paths      & \% of \textit{sources} that initialize with a\\ & path to \textit{home} tile          \\ \hline
Home Center     & 1 if \textit{home} is near center of board, else 0                  \\ \hline
Home Blocks     & 1 if no \textit{blocks} near \textit{home}, else 0 \\ \hline
\end{tabular}
\caption{The constraint factors present in the generator}
\label{table:constraint-factors}
\end{table}

\subsection{Procedure}\label{sec:case-procedure}
\begin{figure*}[tb]
\begin{center}
\includegraphics[width=1.0\linewidth]{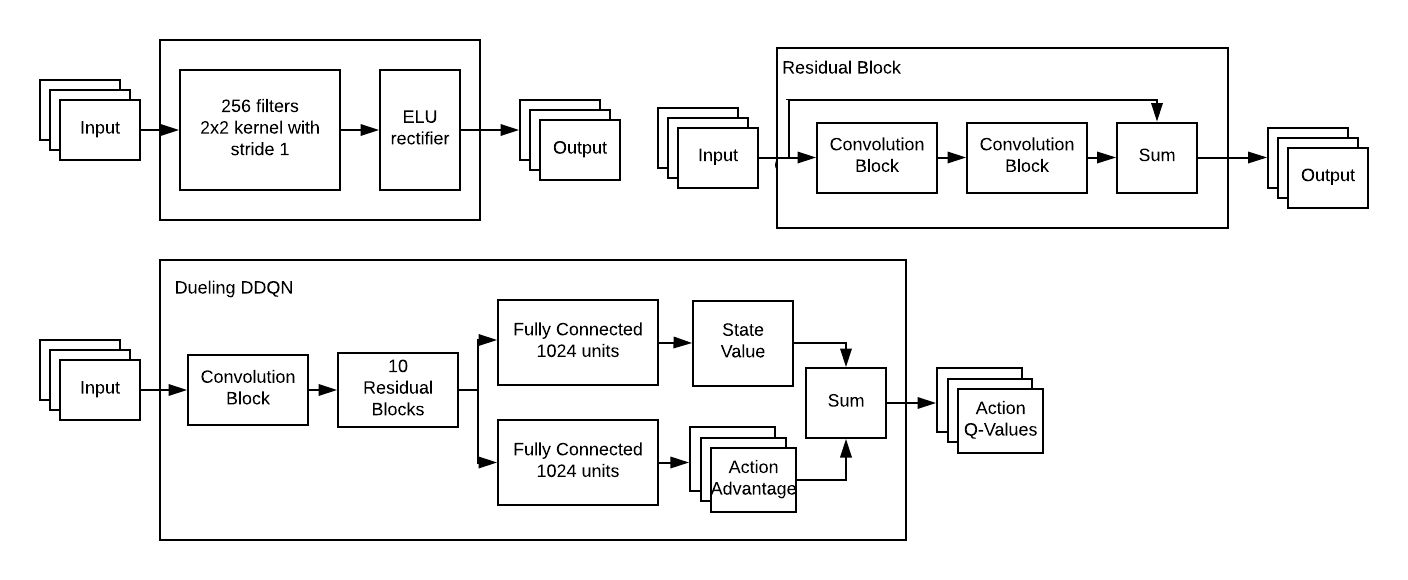}
\caption{The architecture of our DDDQN consists of a convolutional layer followed by a tower of 10 residual blocks. The final residual block is fed to two separate fully-connected layers to produce the current state’s Q-value and the predicted advantage of each possible action. The streams are combined to produce predicted action Q-values.}
\label{fig:network-diagram}
\end{center}
\end{figure*}

\begin{figure*}[tb]
\begin{center}
\includegraphics[width=1.0\linewidth]{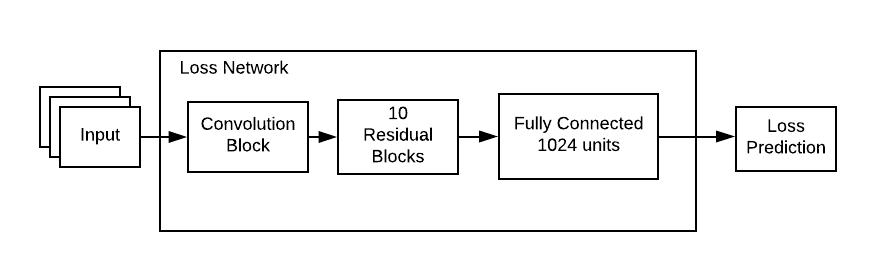}
\caption{The architecture of our loss network consists of a convolutional layer followed by a tower of 10 residual blocks which feeds a fully connected layer that outputs the loss prediction.}
\label{fig:loss-network-diagram}
\end{center}
\end{figure*}

To evaluate the effectiveness of an evolutionarily-based curriculum inside a reinforcement learning training loop, we created several training schedules. A Double Dueling Deep Q Network~\cite{wang2016dueling} (DDDQN) using prioritized replay and a separate loss network was trained from initialization on each schedule. Figure \ref{fig:network-diagram} displays the architecture of this network, and Figure \ref{fig:loss-network-diagram} displays the separate loss network. 

The replay bank of the DDDQN holds $20,000$ training experiences using hyperparameters $\alpha=0.6$, $\beta_{0}=0.4$ annealed to $\beta=1.0$ over the first $1000$ games. To create experiences for the replay bank, the network plays a map from \emph{Attackers and Defenders}, after which it stores all encountered initial states, actions, next states, and rewards as experiences in the bank. The network updates its weights every $5$ maps it plays. A training cycle consists of $250$ batches which contain $32$ samples selected according to prioritized replay. The Q-value update uses a future discount factor $\gamma=0.99$. The loss of each individual experience is used to update the priority. Afterwards, the loss network is trained using the initial state as input and the loss as the target.

\begin{figure}[!htb]
\begin{center}
\includegraphics[width=1.0\linewidth]{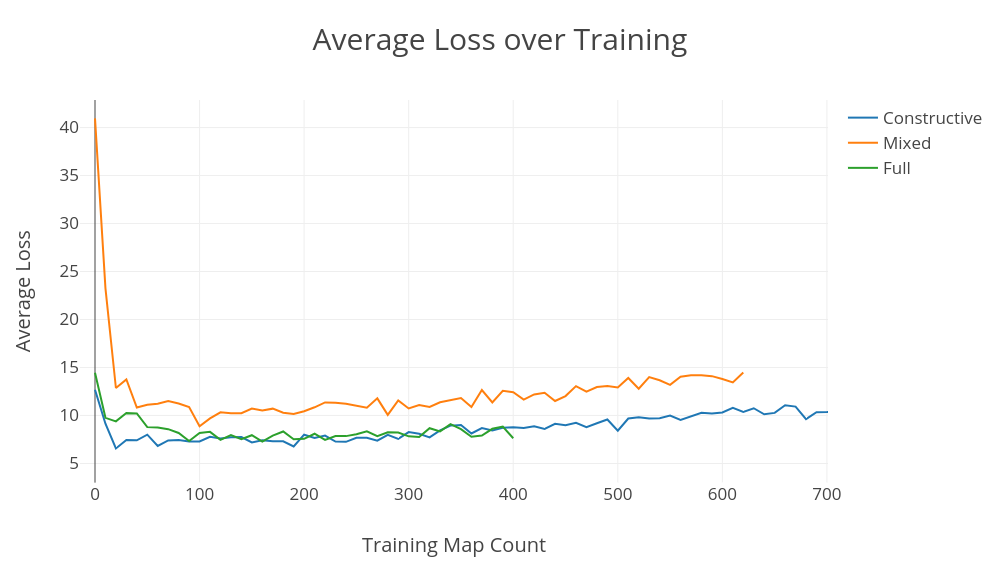}
\caption{The loss over time for the training of each network. Loss was collected every $5$ maps of training by averaging the loss across the $5$ maps.}
\label{fig:loss-results}
\end{center}
\end{figure}

All schedules begin with $50$ maps created using a constructive generator to train the loss network such that it’s output is usable to assess a map’s loss potential. This constructive generator provides an undirected random sampling of the game space. The $50$ starting maps are used are the identical for every schedule. After these initial maps, the schedules differ. The first of these schedules continues to contain only maps constructed by the constructive generator. The second of these schedules contains only evolutionarily-curated maps. The third network contained equal mix of randomly generated maps and evolutionarily-curated maps. Table \ref{table:networks} defines each network's training curriculum.

For each schedule, the network was tested and scored on a fixed set of $1000$ randomly generated maps after every $200$ training maps. The network was optimizing to slay the maximum amount of attackers over the course of play and was scored based on how many of these attackers were slain before an attacker reached the home tile. Training continued until the network failed to improve for two consecutive testing cycles.

\begin{table}[]
\begin{tabular}{|l|l|}
\hline
\textbf{Network} & \textbf{Curriculum} \\ \hline
DQN 1  & 50 + 100\% randomly constructed maps\\ \hline
DQN 2  & 50 + 100\% evolutionarily-curated maps\\ \hline
DQN 3  & 50 + 50\% randomly constructed maps \& \\ 
       & 50\% evolutionarily-curated maps\\ \hline
\end{tabular}
\caption{The networks and their training curriculum ratios}
\label{table:networks}
\end{table}

\begin{figure*}[!htb]
\begin{center}
\includegraphics[width=1.0\linewidth]{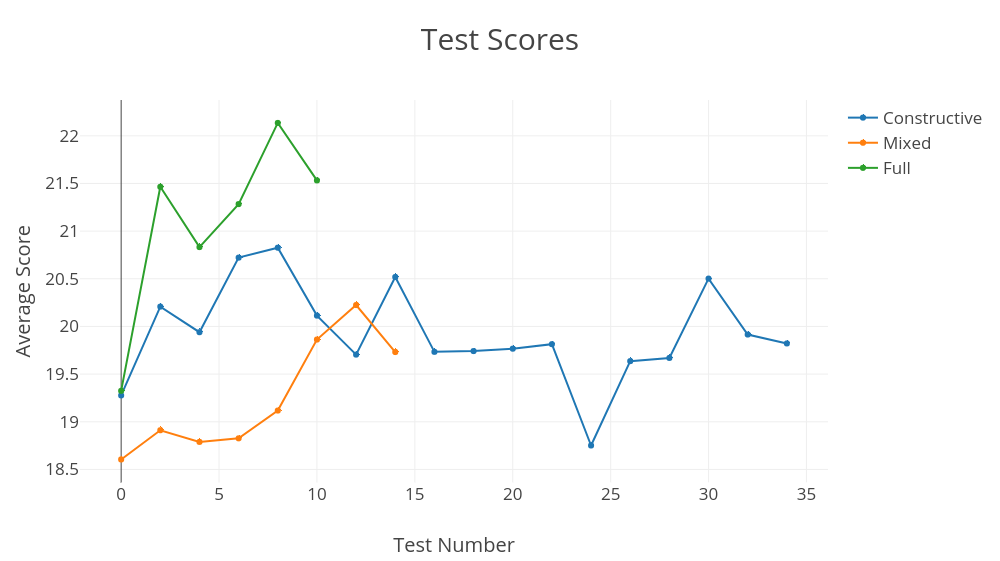}
\caption{The testing score results averaged in $1,000$ map batches over the course of training. The y-intercept marks network performance at network initialization with randomized weights.}
\label{fig:testing-results}
\end{center}
\end{figure*}

\section{Results \& Discussion}\label{sec:results/discussion}
Here, we present the results of the previously described case study. We compare the results of the fully-evolved-curriculum-trained network (full network), the mixed-evolved-curriculum-trained network (mixed network), and the randomly-generated-curriculum-trained (constructive network) trained on randomly sampled maps.

% This section is divided into a testing results comparison (Section \ref{sec:results/discussion-testing}) a training results comparison (Section \ref{sec:results/discussion-training}), and a network loss comparison (Section \ref{sec:results/discussion-loss}).

% \subsection{Network Testing}\label{sec:results/discussion-testing}
As Figure \ref{fig:testing-results} demonstrates, the full network generalizes well within $400$ maps of training with a score of $21.47$, peaking after just $1,600$ maps at $22.14$. In comparison, the constructive network never reaches this score. Even after training on $6,800$ maps, it only peaks at $20.83$ at $1,600$ maps. The mixed network peaked at $20.22$ after $2,000$ maps which is slightly below the constructive network's peak.

Figure \ref{fig:loss-results} displays the loss of each network. Each tick displays the average loss of from a training cycle containing $250$ batches, out to a total of $3,500$ maps. The full network starts out at $14.44$, higher than the constructive network at $12.67$. This suggests that the full network learning was presented with maps which high learning potential, as expected. Very quickly the two networks converge and hover between $7.5$ and $9$ which corresponds the full network's peak performance within the first several hundred maps. The constructive network gradually increases over time to $10$ then stabilizes after it reached peak performance. Contrary to expectations, the mixed network on the other hand shows a much higher loss than either of the other two networks starting at $40.95$ and remaining substantially higher ranging from $10$ to $14$. 

This suggests that the mixed network failed to generalize when presented with an equal mix of evolved maps and generated maps. The network appears to have learned how to discriminate evolve maps from randomly generated maps in a manner that harmed performance on the test set. Specifically, the network learned two classes of strategies: evolved and constructive. As a result, the learning from evolved examples would not generalize correctly to constructive maps that were used in the testing set. By contrast, the full network only saw evolved maps after the first $50$ maps and was able to generalize the strategies learned from evolved maps to randomly generated maps in the test set.

Lastly, the sole requirement of the evolutionary generator is to specify the parameters of the game itself and is generalizable. The network informs the generator of its loss function which makes it not specific to any domain. It is sufficient for the network set up to have the ability to interact with the game  as the architecture is not dependent on it. Since both halves of the systems (generator and network) are generalizable themselves, slight modifications tot he system can make it adapt very well to a new scenario. 

\section{Conclusion}\label{sec:conclusion}
In this paper, we have introduced evolutionarily-curated curriculum learning as a new methodology to train reinforcement agents. We performed a case study using a game we created called \emph{Attackers and Defenders} to prove the validity and effectiveness of this new method. Specifically, we tested a Double Dueling Deep Q-network (DDDQN) with prioritized replay and a separate loss network using this method.

Based on our results, our initial hypothesis that evolutionarily-curated curriculum learning helps networks generalize better and faster than undirected sampling, has been proven true in this environment. Even after nearly three times the amount of training time, the constructive trained network never approaches the performance of the full evolutionarily-curated network. Therefore, it appears that this new training methodology, ECCL, can be used to both expedite training and increase generalization or max performance. 

However, the mixed network does not appear to perform as well despite the fact that its loss values are much higher. This was a result we did not initially expect, allowing for any amount of evolutionarily-based curriculum learning would improve network training. Upon inspection, the mixed network spent considerable effort in differentiating between maps coming from the constructive generator vs. the evolutionary generator. This suggests that a discriminator network trained to predict whether a map was constructed or evolved could be added to the evolutionary generator's fitness functions. Another possibility would be using a similarity metric in fitness to ensure evolved maps are sufficiently different from previously evolved maps to prevent the network from learning to recognize evolved maps.

Given the generalizability of ECCL as in the discussion from the previous section, as it only requires a data generator and a game-playing agent architecture, we also expect it to work well with AlphaGo Zero-based agents as well, and leave that open for future work.

%References and End of Paper%These lines must be placed at the end of your paper
\bibliography{bibliography.bib}

\begin{thebibliography}{}

\bibitem[\protect\citeauthoryear{Anthony, Tian, and
  Barber}{2017}]{anthony2017thinking}
Anthony, T.; Tian, Z.; and Barber, D.
\newblock 2017.
\newblock Thinking fast and slow with deep learning and tree search.
\newblock In {\em Advances in Neural Information Processing Systems},
  5360--5370.

\bibitem[\protect\citeauthoryear{Ashlock, Lee, and
  McGuinness}{2011}]{ashlock2011search}
Ashlock, D.; Lee, C.; and McGuinness, C.
\newblock 2011.
\newblock Search-based procedural generation of maze-like levels.
\newblock {\em IEEE Transactions on Computational Intelligence and AI in Games}
  3(3):260--273.

\bibitem[\protect\citeauthoryear{Ashlock}{2010}]{ashlock2010automatic}
Ashlock, D.
\newblock 2010.
\newblock Automatic generation of game elements via evolution.
\newblock In {\em Computational Intelligence and Games (CIG), 2010 IEEE
  Symposium on},  289--296.
\newblock IEEE.

\bibitem[\protect\citeauthoryear{Ashlock}{2015}]{ashlock2015evolvable}
Ashlock, D.
\newblock 2015.
\newblock Evolvable fashion-based cellular automata for generating cavern
  systems.
\newblock In {\em Computational Intelligence and Games (CIG), 2015 IEEE
  Conference on},  306--313.
\newblock IEEE.

\bibitem[\protect\citeauthoryear{Barto, Sutton, and
  Anderson}{1983}]{barto1983neuronlike}
Barto, A.~G.; Sutton, R.~S.; and Anderson, C.~W.
\newblock 1983.
\newblock Neuronlike adaptive elements that can solve difficult learning
  control problems.
\newblock {\em IEEE transactions on systems, man, and cybernetics}
  (5):834--846.

\bibitem[\protect\citeauthoryear{Bengio \bgroup et al\mbox.\egroup
  }{2009}]{bengio2009curriculum}
Bengio, Y.; Louradour, J.; Collobert, R.; and Weston, J.
\newblock 2009.
\newblock Curriculum learning.
\newblock In {\em Proceedings of the 26th annual international conference on
  machine learning},  41--48.
\newblock ACM.

\bibitem[\protect\citeauthoryear{Cai \bgroup et al\mbox.\egroup
  }{2018}]{cai2018curriculum}
Cai, Q.; Du, M.; Liu, C.; and Song, D.
\newblock 2018.
\newblock Curriculum adversarial training.
\newblock {\em CoRR} abs/1805.04807.

\bibitem[\protect\citeauthoryear{Elman}{1993}]{elman1993learning}
Elman, J.~L.
\newblock 1993.
\newblock Learning and development in neural networks: The importance of
  starting small.
\newblock {\em Cognition} 48(1):71--99.

\bibitem[\protect\citeauthoryear{Floreano, D{\"u}rr, and
  Mattiussi}{2008}]{floreano2008neuroevolution}
Floreano, D.; D{\"u}rr, P.; and Mattiussi, C.
\newblock 2008.
\newblock Neuroevolution: from architectures to learning.
\newblock {\em Evolutionary Intelligence} 1(1):47--62.

\bibitem[\protect\citeauthoryear{Green \bgroup et al\mbox.\egroup
  }{2018}]{green2018generating}
Green, M.~C.; Khalifa, A.; Barros, G.~A.; Nealen, A.; and Togelius, J.
\newblock 2018.
\newblock Generating levels that teach mechanics.
\newblock In {\em Proceedings of the Foundation of Digital Games}.

\bibitem[\protect\citeauthoryear{Gruau and others}{1994}]{gruau1994neural}
Gruau, F., et~al.
\newblock 1994.
\newblock Neural network synthesis using cellular encoding and the genetic
  algorithm.

\bibitem[\protect\citeauthoryear{Gullapalli}{1990}]{gullapalli1990stochastic}
Gullapalli, V.
\newblock 1990.
\newblock A stochastic reinforcement learning algorithm for learning
  real-valued functions.
\newblock {\em Neural networks} 3(6):671--692.

\bibitem[\protect\citeauthoryear{Hessel \bgroup et al\mbox.\egroup
  }{2017}]{hessel2017rainbow}
Hessel, M.; Modayil, J.; Van~Hasselt, H.; Schaul, T.; Ostrovski, G.; Dabney,
  W.; Horgan, D.; Piot, B.; Azar, M.; and Silver, D.
\newblock 2017.
\newblock Rainbow: Combining improvements in deep reinforcement learning.
\newblock {\em arXiv preprint arXiv:1710.02298}.

\bibitem[\protect\citeauthoryear{Khalifa and
  Fayek}{2015a}]{khalifa2015automatic}
Khalifa, A., and Fayek, M.
\newblock 2015a.
\newblock Automatic puzzle level generation: A general approach using a
  description language.
\newblock In {\em Computational Creativity and Games Workshop}.

\bibitem[\protect\citeauthoryear{Khalifa and
  Fayek}{2015b}]{khalifa2015literature}
Khalifa, A., and Fayek, M.
\newblock 2015b.
\newblock Literature review of procedural content generation in puzzle games.
\newblock \url{http://www.akhalifa.com/documents/LiteratureReviewPCG.pdf}.

\bibitem[\protect\citeauthoryear{Khalifa \bgroup et al\mbox.\egroup
  }{2016}]{khalifa2016general}
Khalifa, A.; Perez-Liebana, D.; Lucas, S.~M.; and Togelius, J.
\newblock 2016.
\newblock General video game level generation.
\newblock In {\em Proceedings of the Genetic and Evolutionary Computation
  Conference 2016},  253--259.
\newblock ACM.

\bibitem[\protect\citeauthoryear{Khalifa \bgroup et al\mbox.\egroup
  }{2017}]{khalifa2017general}
Khalifa, A.; Green, M.~C.; Perez-Liebana, D.; and Togelius, J.
\newblock 2017.
\newblock General video game rule generation.
\newblock In {\em Computational Intelligence and Games (CIG), 2017 IEEE
  Conference on},  170--177.
\newblock IEEE.

\bibitem[\protect\citeauthoryear{Khalifa \bgroup et al\mbox.\egroup
  }{2018}]{khalifa2018talakat}
Khalifa, A.; Lee, S.; Nealen, A.; and Togelius, J.
\newblock 2018.
\newblock Talakat: Bullet hell generation through constrained map-elites.
\newblock In {\em Proceedings of The Genetic and Evolutionary Computation
  Conference}.
\newblock ACM.

\bibitem[\protect\citeauthoryear{Kimbrough \bgroup et al\mbox.\egroup
  }{2008}]{kimbrough2008feasible}
Kimbrough, S.~O.; Koehler, G.~J.; Lu, M.; and Wood, D.~H.
\newblock 2008.
\newblock On a feasible--infeasible two-population (fi-2pop) genetic algorithm
  for constrained optimization: Distance tracing and no free lunch.
\newblock {\em European Journal of Operational Research} 190(2):310--327.

\bibitem[\protect\citeauthoryear{Krueger and Dayan}{2009}]{krueger2009flexible}
Krueger, K.~A., and Dayan, P.
\newblock 2009.
\newblock Flexible shaping: How learning in small steps helps.
\newblock {\em Cognition} 110(3):380--394.

\bibitem[\protect\citeauthoryear{McGuinness and
  Ashlock}{2011}]{mcguinness2011decomposing}
McGuinness, C., and Ashlock, D.
\newblock 2011.
\newblock Decomposing the level generation problem with tiles.
\newblock In {\em Evolutionary Computation (CEC), 2011 IEEE Congress on},
  849--856.
\newblock IEEE.

\bibitem[\protect\citeauthoryear{Miller, Thomson, and
  Fogarty}{1997}]{miller1997designing}
Miller, J.~F.; Thomson, P.; and Fogarty, T.
\newblock 1997.
\newblock Designing electronic circuits using evolutionary algorithms.
  arithmetic circuits: A case study.

\bibitem[\protect\citeauthoryear{Minsky}{1954}]{minsky1954theory}
Minsky, M.~L.
\newblock 1954.
\newblock {\em Theory of neural-analog reinforcement systems and its
  application to the brain model problem}.
\newblock Princeton University.

\bibitem[\protect\citeauthoryear{Mnih \bgroup et al\mbox.\egroup
  }{2015}]{mnih2015human}
Mnih, V.; Kavukcuoglu, K.; Silver, D.; Rusu, A.~A.; Veness, J.; Bellemare,
  M.~G.; Graves, A.; Riedmiller, M.; Fidjeland, A.~K.; Ostrovski, G.; et~al.
\newblock 2015.
\newblock Human-level control through deep reinforcement learning.
\newblock {\em Nature} 518(7540):529.

\bibitem[\protect\citeauthoryear{Risi and
  Togelius}{2017}]{risi2017neuroevolution}
Risi, S., and Togelius, J.
\newblock 2017.
\newblock Neuroevolution in games: State of the art and open challenges.
\newblock {\em IEEE Transactions on Computational Intelligence and AI in Games}
  9(1):25--41.

\bibitem[\protect\citeauthoryear{Ronald and
  Schoenauer}{1994}]{ronald1994genetic}
Ronald, E., and Schoenauer, M.
\newblock 1994.
\newblock Genetic lander: An experiment in accurate neuro-genetic control.
\newblock In {\em International Conference on Parallel Problem Solving from
  Nature},  452--461.
\newblock Springer.

\bibitem[\protect\citeauthoryear{Rumelhart, Hinton, and
  Williams}{1986}]{rumelhart1986learning}
Rumelhart, D.~E.; Hinton, G.~E.; and Williams, R.~J.
\newblock 1986.
\newblock Learning representations by back-propagating errors.
\newblock {\em nature} 323(6088):533.

\bibitem[\protect\citeauthoryear{Schaul \bgroup et al\mbox.\egroup
  }{2015}]{schaul2015prioritized}
Schaul, T.; Quan, J.; Antonoglou, I.; and Silver, D.
\newblock 2015.
\newblock Prioritized experience replay.
\newblock {\em arXiv preprint arXiv:1511.05952}.

\bibitem[\protect\citeauthoryear{Schmidhuber}{2015}]{schmidhuber2015deep}
Schmidhuber, J.
\newblock 2015.
\newblock Deep learning in neural networks: An overview.
\newblock {\em Neural networks} 61:85--117.

\bibitem[\protect\citeauthoryear{Silver \bgroup et al\mbox.\egroup
  }{2016}]{silver2016mastering}
Silver, D.; Huang, A.; Maddison, C.~J.; Guez, A.; Sifre, L.; Van Den~Driessche,
  G.; Schrittwieser, J.; Antonoglou, I.; Panneershelvam, V.; Lanctot, M.;
  et~al.
\newblock 2016.
\newblock Mastering the game of go with deep neural networks and tree search.
\newblock {\em nature} 529(7587):484.

\bibitem[\protect\citeauthoryear{Silver \bgroup et al\mbox.\egroup
  }{2017a}]{silver2017masteringchess}
Silver, D.; Hubert, T.; Schrittwieser, J.; Antonoglou, I.; Lai, M.; Guez, A.;
  Lanctot, M.; Sifre, L.; Kumaran, D.; Graepel, T.; et~al.
\newblock 2017a.
\newblock Mastering chess and shogi by self-play with a general reinforcement
  learning algorithm.
\newblock {\em arXiv preprint arXiv:1712.01815}.

\bibitem[\protect\citeauthoryear{Silver \bgroup et al\mbox.\egroup
  }{2017b}]{silver2017mastering}
Silver, D.; Schrittwieser, J.; Simonyan, K.; Antonoglou, I.; Huang, A.; Guez,
  A.; Hubert, T.; Baker, L.; Lai, M.; Bolton, A.; et~al.
\newblock 2017b.
\newblock Mastering the game of go without human knowledge.
\newblock {\em Nature} 550(7676):354.

\bibitem[\protect\citeauthoryear{Sutton and
  Barto}{1998}]{sutton1998introduction}
Sutton, R.~S., and Barto, A.~G.
\newblock 1998.
\newblock {\em Introduction to reinforcement learning}, volume 135.
\newblock MIT press Cambridge.

\bibitem[\protect\citeauthoryear{Sutton}{1984}]{sutton1984temporal}
Sutton, R.~S.
\newblock 1984.
\newblock Temporal credit assignment in reinforcement learning.

\bibitem[\protect\citeauthoryear{Sutton}{1988}]{sutton1988learning}
Sutton, R.~S.
\newblock 1988.
\newblock Learning to predict by the methods of temporal differences.
\newblock {\em Machine learning} 3(1):9--44.

\bibitem[\protect\citeauthoryear{Szepesv{\'a}ri}{2010}]{szepesvari2010algorithms}
Szepesv{\'a}ri, C.
\newblock 2010.
\newblock Algorithms for reinforcement learning.
\newblock {\em Synthesis lectures on artificial intelligence and machine
  learning} 4(1):1--103.

\bibitem[\protect\citeauthoryear{Togelius, Shaker, and
  Nelson}{2016}]{togelius2016introduction}
Togelius, J.; Shaker, N.; and Nelson, M.~J.
\newblock 2016.
\newblock The search-based approach.
\newblock In Shaker, N.; Togelius, J.; and Nelson, M.~J., eds., {\em Procedural
  Content Generation in Games: A Textbook and an Overview of Current Research}.
  Springer.
\newblock  17--30.

\bibitem[\protect\citeauthoryear{Van~Hasselt, Guez, and
  Silver}{2016}]{van2016deep}
Van~Hasselt, H.; Guez, A.; and Silver, D.
\newblock 2016.
\newblock Deep reinforcement learning with double q-learning.
\newblock In {\em AAAI}, volume~2, ~5.
\newblock Phoenix, AZ.

\bibitem[\protect\citeauthoryear{Wang \bgroup et al\mbox.\egroup
  }{2016}]{wang2016dueling}
Wang, Z.; Schaul, T.; Hessel, M.; Hasselt, H.; Lanctot, M.; and Freitas, N.
\newblock 2016.
\newblock Dueling network architectures for deep reinforcement learning.
\newblock In {\em International Conference on Machine Learning},  1995--2003.

\bibitem[\protect\citeauthoryear{Watkins and Dayan}{1992}]{watkins1992q}
Watkins, C.~J., and Dayan, P.
\newblock 1992.
\newblock Q-learning.
\newblock {\em Machine learning} 8(3-4):279--292.

\bibitem[\protect\citeauthoryear{Williams}{1992}]{williams1992simple}
Williams, R.~J.
\newblock 1992.
\newblock Simple statistical gradient-following algorithms for connectionist
  reinforcement learning.
\newblock {\em Machine learning} 8(3-4):229--256.

\end{thebibliography}
\bibliographystyle{aaai}
\end{document}